\documentclass[10pt,twocolumn,letterpaper]{article}

\usepackage[pagenumbers]{cvpr} 

\usepackage{graphicx}
\usepackage{amsmath}
\usepackage{amssymb}
\usepackage{booktabs}
\usepackage[accsupp]{axessibility}  

\usepackage[pagebackref,breaklinks]{hyperref}

\usepackage[capitalize]{cleveref}
\usepackage{multirow}

\newcommand{\mypara}[1]{\vspace{2pt}\noindent{\bf{#1}}}
\newcommand{\modelName}{AVCA\xspace}
\def\cvprPaperID{10849} 
\def\confName{CVPR}
\def\confYear{2022}

\begin{document}

\title{Audio-visual Generalised Zero-shot Learning \\ with Cross-modal Attention and Language}


\author{Otniel-Bogdan Mercea\textsuperscript{1}, \hspace{2pt} Lukas Riesch\textsuperscript{1,2}, \hspace{2pt} A. Sophia Koepke\textsuperscript{1}, \hspace{2pt} 
Zeynep Akata\textsuperscript{1,3,4}  \\ \\
{\textsuperscript{1}University of T{\"u}bingen \hspace{2pt}
\textsuperscript{2}Robert Bosch GmbH \hspace{2pt}
\textsuperscript{3}MPI for Informatics}\\
{\textsuperscript{4}MPI for Intelligent Systems} \\
{\small \tt \{otniel-bogdan.mercea, a-sophia.koepke, zeynep.akata\}@uni-tuebingen.de} \\
{\small \tt lukas.riesch@de.bosch.com}
}
\maketitle

\begin{abstract}
    Learning to classify video data from classes not included in the training data, i.e. video-based zero-shot learning, is challenging. We conjecture that the natural alignment between the audio and visual modalities in video data provides a rich training signal for learning discriminative multi-modal representations. Focusing on the relatively underexplored task of audio-visual zero-shot learning, we propose to learn multi-modal representations from audio-visual data using cross-modal attention and exploit textual label embeddings for transferring knowledge from seen classes to unseen classes. Taking this one step further, in our generalised audio-visual zero-shot learning setting,  we include all the training classes in the test-time search space which act as distractors and increase the difficulty while making the setting more realistic. Due to the lack of a unified benchmark in this domain, we introduce a (generalised) zero-shot learning benchmark on three audio-visual datasets of varying sizes and difficulty, VGGSound, UCF, and ActivityNet, ensuring that the unseen test classes do not appear in the dataset used for supervised training of the backbone deep models. Comparing multiple relevant and recent methods, we demonstrate that our proposed AVCA model achieves state-of-the-art performance on all three datasets. Code and data are available at \url{https://github.com/ExplainableML/AVCA-GZSL}.
\end{abstract}

\section{Introduction}\label{sec:intro}

\begin{figure}[t]
    \centering
    \includegraphics[width=0.96\columnwidth]{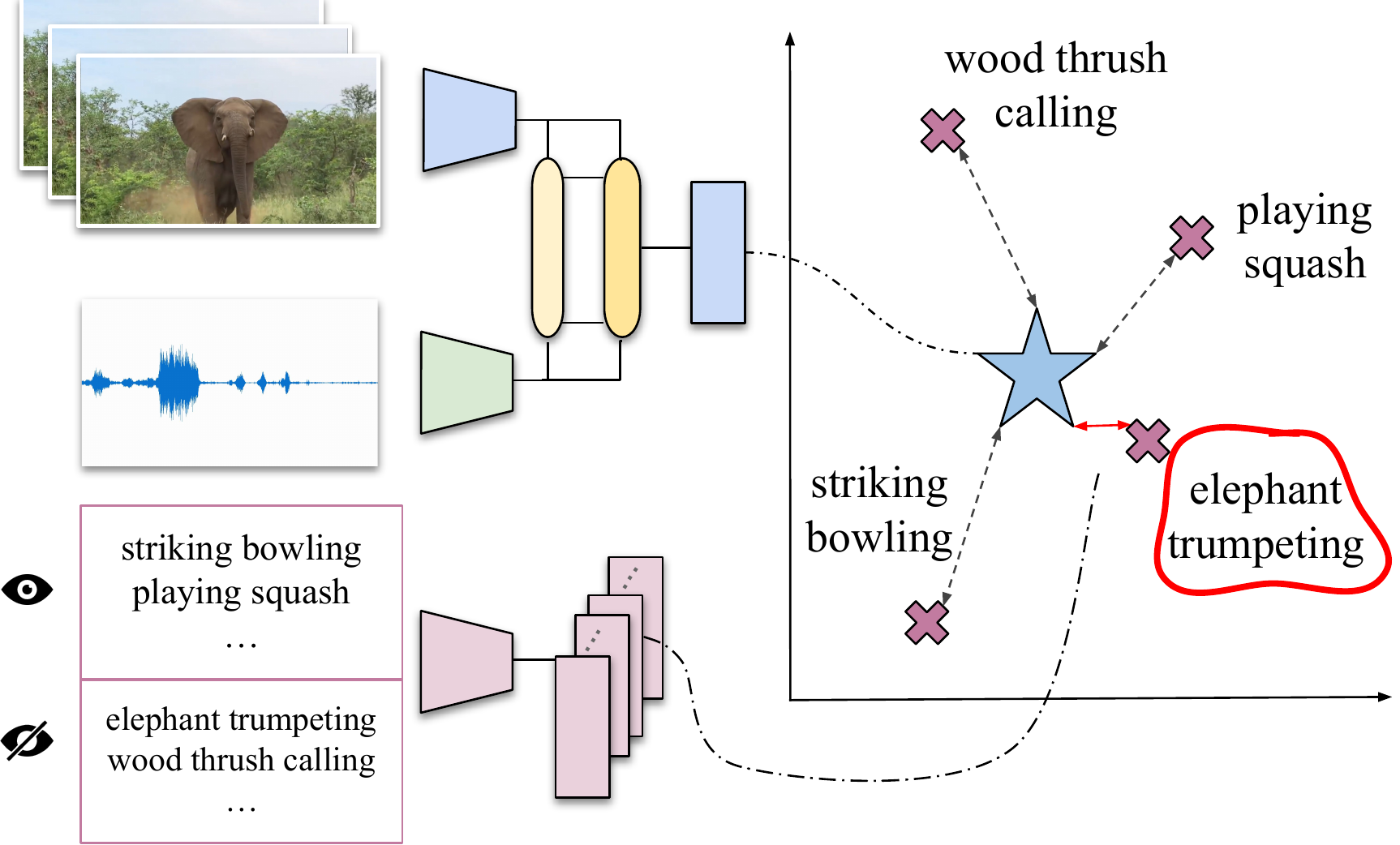}
    \caption{Our audio-visual (generalised) ZSL framework aligns an audio-visual embedding with the corresponding textual label embedding via cross-modal attention. 
    It can classify videos from previously unseen classes (\eg \textit{elephant trumpeting}) by predicting the class (red) whose textual label embedding (purple cross) is closest to the audio-visual embedding (blue star).}
    \label{fig:teaser}
    \vspace{-0.7em}
\end{figure}

Most zero-shot learning (ZSL) methods developed for image classification ~\cite{verma2018generalized,schonfeld2019generalized,xu2020attribute, akata2015evaluation, akata2015label, romera2015embarrassingly} and action recognition~\cite{brattoli2020rethinking, bishay2019tarn, hahn2019action2vec, xu2016multi} only use unimodal input, e.g. images. 
However, humans leverage multi-modal sensory inputs in their everyday activities. Imagine the situation in which the sound of a dog barking is audible but the dog is visually occluded. In this case, we cannot understand the scene when relying on visual information alone.
Using multiple modalities, such as vision and sound, allows to gather context and capture complementary information.
Similarly, using both visual and audio information allows for a richer training signal for learning frameworks. 
This paper investigates the challenging task of (generalised) ZSL with multi-modal audio-visual data by leveraging the natural alignment of audio and visual information in videos.

Recently, \cite{parida2020coordinated, mazumder2021avgzslnet} have explored the task of zero-shot video recognition using multi-modal visual and audio information as inputs. However, the AudioSetZSL dataset~\cite{parida2020coordinated} used for this, contains an overlap between the classes used for validation and testing. This results in learning stronger representations for classes overlapping with the training and validation sets (which covers all the classes in this dataset) and hinders the model's capability to learn sufficiently generalisable representations that allow information transfer. In real-world applications, such models perform well on seen classes, but poorly on previously truly unseen classes.
In this work, we propose three benchmarks of varying size and difficulty curated from the \mbox{VGG}Sound~\cite{chen2020vggsound}, UCF101~\cite{soomro2012ucf101}, and ActivityNet~\cite{caba2015activitynet} datasets that could act as a unified and challenging playground for Generalised ZSL (GZSL) and ZSL research in the audio-visual domain. We suggest using audio and visual features extracted using SeLaVi~\cite{asano2020labelling} pretrained using self-supervision. Throughout this work, we use features that were obtained from training in a self-supervised fashion to reduce the information leakage from supervised pre-training to the zero-shot task which has been identified as a problem in other ZSL benchmarks~\cite{brattoli2020rethinking}.

We tackle the audio-visual generalised zero-shot learning task with our Audio-Visual Cross-Attention (\modelName) framework which is trained to align a rich learnt audio-visual representation with textual label embeddings. Our multi-stream architecture contains an audio and a visual branch which exchange information using cross-attention between the two modalities. \modelName is computationally lightweight and efficient since it uses audio and visual features extracted from pretrained networks as inputs instead of raw audio and image data.
Our proposed framework is trained using multiple novel loss functions that are based on triplet losses and a regularisation loss that ensures that salient unimodal information is preserved in the learnt multi-modal representations. 
Our experiments show that \modelName achieves state-of-the-art performance on the three introduced benchmark datasets. We show that using multi-modal input data leads to stronger (G)ZSL performance than using unimodal data. 

To summarise, our contributions are as follows:
(1) We introduce three novel benchmarks for audio-visual (generalised) zero-shot learning curated from the VGGSound, UCF101, and ActivityNet datasets;
(2) We propose \modelName, a cross-modal model for audio-visual (G)ZSL which leverages cross-modal attention between audio and visual information; 
(3) We show that \modelName yields state-of-the-art performance on all proposed audio-visual (G)ZSL benchmarks, outperforming the state-of-the-art unimodal and multi-modal zero-shot learning methods. Furthermore, we provide a qualitative analysis of the learnt multi-modal embedding space, demonstrating well-separated clustering for both seen and unseen classes.

\section{Related Work}\label{sec:related}
We review audio-visual learning, ZSL with image, video and audio data, and audio-visual ZSL. 

\mypara{Audio-visual learning.}
Audio-visual learning has enabled tremendous progress for numerous applications, such as for separating and localising sounds in videos~\cite{owens2018audio,tian2018audio,arandjelovic2018objects,gao2019co,chen2021localizing,Afouras20b,afouras2021selfsupervised,qian2020multiple,xu2020cross,tzinis2020into,zhu2022v,zhao2018sound,zhao2019sound}, audio-visual synchronisation~\cite{chen2021audio,chung2016out,ebeneze2021detection,khosravan2019attention}, person-clustering in videos~\cite{brown2021face}, (visual) speech and speaker recognition~\cite{afouras2020asr, afouras2018deep,nagrani2020disentangled}, spotting of spoken keywords~\cite{momeni2020seeing,Prajwal2021}, audio synthesis using visual information~\cite{zhou2019vision, goldstein2018guitar,su2020multi,gan2020foley,narasimhan2021strumming,koepke2020sight,koepke2019visual,su2021does}, and audio-driven image synthesis~\cite{wiles2018x2face,jamaludin2019you}.
Additionally, the natural alignment between audio and visual data in videos has been leveraged to learn powerful audio-visual representations for video or audio classification~\cite{owens2016ambient,owens2018learning,alwassel2019self,patrick2020multi,korbar2018cooperative,aytar2016soundnet,chen2021distilling,asano2020labelling,nagrani2021attention,xiao2020audiovisual,cheng2020look}.
In contrast to those methods, we consider the ZSL setting for classification.

\mypara{ZSL with images, videos and audio.}
Recently, numerous image-based generative ZSL methods have been proposed
~\cite{xian2019f,narayan2020latent,zhu2019learning,xian2018feature,zhu2018generative,verma2018generalized,schonfeld2019generalized}.
Their drawback is that the unseen classes need to be known a priori. In contrast, non-generative methods~\cite{xu2020attribute, akata2015evaluation, akata2015label, romera2015embarrassingly, kodirov2017semantic, xian2018zero, frome2013devise,xu2022vgse} learn a mapping from input features to semantics of the classes (\eg textual class label embeddings). Our \modelName model also learns to map its inputs to textual embeddings, but it leverages cross-attention between the audio and visual input modalities rather than using only visual inputs.

Video-based ZSL has been addressed by multiple recent works~\cite{brattoli2020rethinking, bishay2019tarn, hahn2019action2vec, xu2016multi, wang2017zero,roitberg2018towards,gowda2021new}. Using features extracted from pretrained networks results in computationally more feasible frameworks~\cite{wang2017zero,hahn2019action2vec,bishay2019tarn} than training end-to-end~\cite{brattoli2020rethinking}. Our model also takes pre-extracted audio and visual features as inputs, resulting in a computationally efficient framework. In order to consider a pure ZSL setting when using pre-extracted features, the overlap between classes used for supervised pre-training of the feature extractors and unseen classes has to be removed~\cite{brattoli2020rethinking,roitberg2018towards,gowda2021new}. This was not done in some of the previous works (\eg~\cite{zhu2018towards,wang2017zero,hahn2019action2vec,bishay2019tarn}). In contrast, we propose three benchmarks for audio-visual (G)ZSL on multi-modal audio-visual video datasets with no overlap between classes used for supervised pre-training and unseen classes.

\begin{figure*}[t]
    \centering
    \includegraphics[width=0.93\linewidth,trim=20 10 0 0]{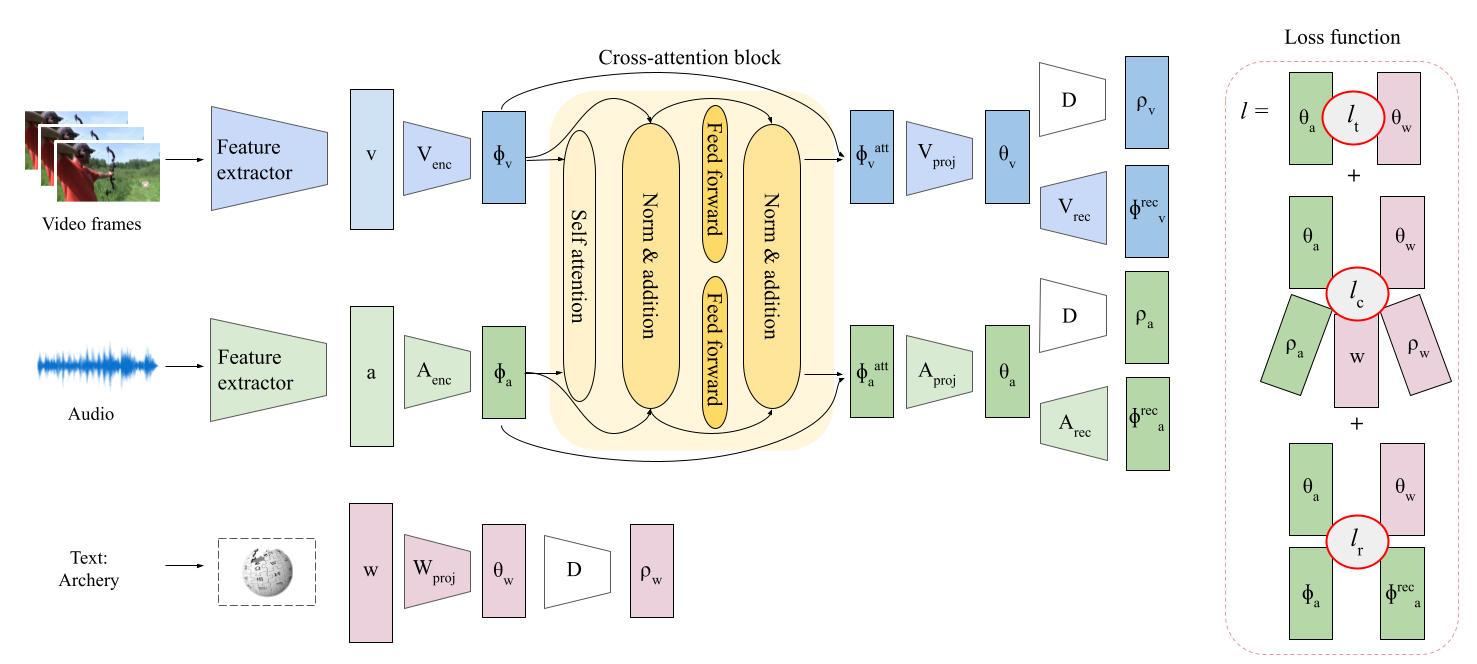}
    \caption{Our Audio-Visual Cross Attention (\modelName) model
    takes visual and audio features as inputs. A cross-attention block allows the sharing of information across modalities. The outputs of the two model branches are trained to be aligned with their corresponding textual label embedding using losses illustrated on the right-hand side. Negative samples for the contrastive loss functions are obtained using visual and audio inputs from different videos which do not share semantic information.
    We only show losses that involve the audio branch, those for the visual branch are similar. At test time, the class prediction is obtained by determining the class for which $\theta_w$ is closest to $\theta_v$.
    }
    \label{fig:model}
    \vspace{-1em}
\end{figure*}

 Methods for zero-shot audio classification~\cite{xie2021zero,xie2021zeroieee} also used textual sound class embeddings (\eg word2vec~\cite{mikolov2013efficient}, BERT~\cite{devlin2018bert}, or GloVe~\cite{pennington2014glove}) or descriptions. \cite{choi2019zero} investigate zero-shot music classification and tagging with word2vec embeddings and human-labeled attribute information (e.g.\ the presence or absence of musical instruments). For our \modelName model, we do not use any attribute information, but instead leverage the semantic alignment between audio and visual information in addition to textual label embeddings.

\mypara{Audio-visual ZSL.} Recently, \cite{parida2020coordinated, mazumder2021avgzslnet} proposed frameworks that consider the task of GZSL from audio-visual data. AVGZSLNet~\cite{mazumder2021avgzslnet} uses late fusion on the AudioSetZSL dataset~\cite{parida2020coordinated} to combine information from the two modalities. Instead, and also different to other audio-visual frameworks~\cite{tian2020unified, xuan2020cross} that use a simple dot-product operation for cross-attention, we use a transformer-based cross-attention mechanism. This allows for early and efficient sharing of multi-modal information, which is further encouraged by our proposed loss functions. Furthermore, the AudioSetZSL dataset~\cite{parida2020coordinated} does not include a validation split with unseen validation classes. Hence, \cite{parida2020coordinated,mazumder2021avgzslnet} select the GZSL hyperparameters directly on the (unseen) test classes. Furthermore, the AudioSetZSL dataset is comparatively small; it uses only 10 test classes as unseen classes. To allow for evaluation of audio-visual ZSL at larger scale and in a pure GZSL setting, we propose new benchmarks on three different audio-visual video datasets. Our proposed benchmarks are suitable for both the GZSL and ZSL tasks.

\section{Audio-Visual Cross Attention (\modelName)}
The goal of audio-visual ZSL from video data is to learn to recognise videos from unseen classes (U), \ie classes that were not seen during training. In the GZSL setting, the test set contains not only samples from unseen classes, but also from seen classes (S). This makes GZSL more challenging and more closely aligned with real-world learning tasks. 

More formally, we denote the training set consisting only of samples from seen classes by $S=(v^s_i,a^s_i,y^s_i)_{i \in \{1,\cdots,N\}}$, where $v^s_i,a^s_i$ are visual and audio features respectively, $y^s_i$ is the corresponding ground-truth class $j$, and $N$ is the number of samples in the training set. We refer to the class-level text embedding for class $j$ as $w^s_j$. The goal is to learn a function $h: (v^s_i, a^s_i) \mapsto w^s_j$ which can then also be applied to samples from unseen classes $h(v^u_i, a^u_i)=w^u_j$, where $(v^u_j,a^u_j, y^u_j) \in U$ for the set of test samples from unseen classes $U=(v^u_i,a^u_i,y^u_i)_{i \in \{1,\cdots,M\}}$ with $M$ samples.

\subsection{Model Architecture}\label{sec:model_arch}
Our \modelName model architecture is visualised in \cref{fig:model}. For easier readability, we dropped the subscripts $i$, $j$, indicating the $i$-th dataset sample and the ground-truth class $j$.

\modelName takes audio and visual features $a, v \in \mathbb{R}^{k_{input}}$ as inputs which are extracted using pretrained feature extractors. Those are passed through two different encoder blocks $A_{enc}$ and $V_{enc}$ for the audio and visual modality respectively, giving embeddings 
\begin{equation}
    A_{enc}(a)= \phi_a \text{ and } V_{enc}(v) = \phi_v
\end{equation} 
with $\phi_a,\phi_v \in \mathbb{R}^{k_{f}}$. 
The encoder blocks each consist of a sequence of two linear layers $f^m_1, f^m_2$ for $m \in \{a,v\}$, where $f^m_1:\mathbb{R}^{k_{input}} \rightarrow \mathbb{R}^{k_{fhidd}}$ and $f^m_2:\mathbb{R}^{k_{fhidd}} \rightarrow \mathbb{R}^{k_{f}}$. $f^m_1, f^m_2$ are each followed by batch normalisation\cite{ioffe2015batch}, a ReLU~\cite{nair2010rectified}, and dropout~\cite{JMLR:v15:srivastava14a} with dropout rate $r_{enc}$.

\mypara{Cross-attention block.}
We propose to use a cross-attention block to share information between the audio and visual representations $\phi_a$ and $\phi_v$. It consists of a multi-head self-attention layer, followed by a fully-connected feed-forward block. Similar to~\cite{vaswani2017attention}, we use a residual connection for the two layers, followed by layer normalisation~\cite{ba2016layer}. 

The feed-forward blocks for the audio and visual branch each consist of a linear projection layer $f_3^m:\mathbb{R}^{k_{f}} \rightarrow \mathbb{R}^{k_{attnhidd}}$ for $m \in \{a,v\}$, followed by GELU~\cite{hendrycks2016gaussian}, dropout with dropout rate of $r_{enc}$, another linear projection layer $f^m_4:\mathbb{R}^{k_{attnhidd}} \rightarrow \mathbb{R}^{k_{f}}$ for $m \in \{a,v\}$ and finally a dropout with dropout rate of $r_{enc}$. The outputs of the cross-attention block are
$\phi_a^{att}, \phi_v^{att} \in \mathbb{R}^{k_{f}}$.

A residual connection around the cross-attention block and subsequent projection blocks $A_{proj}$ and $V_{proj}$ give
\begin{equation}
A_{proj}(\phi_a^{att} + \phi_a)=\theta_a \text{ and } V_{proj}(\phi_v^{att} + \phi_v)=\theta_v,
\end{equation}
where $\theta_a, \theta_v \in \mathbb{R}^{k_{proj}}$. The projection blocks each consist of a sequence of two linear layers $f^m_5$ and $f^m_6$ for $m \in \{a,v\}$, where $f^m_5:\mathbb{R}^{k_{f}} \rightarrow \mathbb{R}^{k_{fhidd}}$ and $f^m_6:\mathbb{R}^{k_{fhidd}} \rightarrow \mathbb{R}^{k_{proj}}$. $f^m_5, f^m_6$ are each followed by batch normalisation, a ReLU, and dropout with dropout rate $r_{proj}$.

Furthermore, the word2vec class label embeddings $w^j$ for class $j$ are passed through the projection block $W_{proj}(w^j)=\theta^j_w$, where $\theta^j_w \in \mathbb{R}^{k_{proj}}$ (in \cref{fig:model} shown without the superscript $j$). $W_{proj}$ consists of a sequence of one linear projection layer, batch normalisation, ReLU, and dropout with dropout rate $r_{dec}$.

At test time, class predictions $c$ are obtained by determining the class $c$ that corresponds to the textual class label embedding that is closest to the multi-modal representation $\theta_v$ (in our experiments we found that using $\theta_a$ gave slightly weaker results):
\begin{equation}
    c = \underset{j}{\mathrm{argmin}}(\| \theta^j_w - \theta_v \|_2).
\end{equation}

\subsection{Loss Functions}\label{sec:losses}
We train our \modelName model using a loss function $l$ consisting of a base triplet loss $l_t$, a composite triplet and reconstruction loss $l_c$, and a regularisation loss $l_r$: 
\begin{equation}\label{eq:full_loss}
    l = l_t + l_c + l_r.
\end{equation}

We use the triplet loss function $t(a,p,n)=\max(\| a -  p\|_2 - \| a - n\|_2 + \mu)$, where $a$ is the anchor embedding, $p$ and $n$ are embeddings for positive samples and negative samples respectively, and $\mu$ is the margin hyperparameter. For triplet losses, we use the superscript $+$ to denote positive samples that match the anchor and $-$ for negative samples that do not semantically match the anchor. For all other losses, we only use matching pairs.

\mypara{Base triplet loss.}
In our base triplet loss $l_{t}$:
\begin{equation}\label{eq:base_triplet}
\begin{aligned}
  l_{t}=t( \theta^{+}_{a}, \theta^{+}_{w}, \theta^{-}_{a})
           +t( \theta^{+}_{v}, \theta^{+}_{w}, \theta^{-}_{v})\\
           +t( \theta^{+}_{w}, \theta^{+}_{a}, \theta^{-}_{w})
           +t( \theta^{+}_{w}, \theta^{+}_{v}, \theta^{-}_{w}),
 \end{aligned}
\end{equation}
where $\theta_m^+$ and $\theta_m^-$ correspond to positive and negative samples respectively for $m \in \{a,v,w\}$, ensuring that the projected visual and audio features $\theta_v$ and $\theta_a$ are aligned with the projected textual features $\theta_w$. This is essential, since at test time, the proximity of $\theta_v$ (which, despite being the output of the visual branch of \modelName, is a multi-modal embedding containing both audio and visual information) to $\theta_w$ for different classes is used to determine the output class.

\mypara{Composite triplet and reconstruction loss.}
Inspired by \cite{mazumder2021avgzslnet}, we additionally use a composite triplet and reconstruction loss and explain its components in more detail below:
\begin{equation}\label{eq:avgzsl_loss}
    l_{c}=l_{rec}+l_{ct}+l_{w}.
\end{equation}
We use a decoder $D: \mathbb{R}^{k_{proj}} \mapsto \mathbb{R}^{k_{w2v}}$, such that $D(\theta_{m}) = \rho_m$ for $m \in \{a,v,w\}$. $D$ consists of a sequence of one linear projection layer, batch normalisation, a ReLU, and dropout with dropout rate $r_{dec}$.
We employ the mean squared error metric $d(b,c)=\frac{1}{n}\sum_{i=1}^n(b_i-c_i)^2$.
The reconstruction loss $l_{rec}$ can then be written as:
\begin{equation}\label{eq:pareto_mle2}
  l_{rec}=d(\rho_a, w) +d(\rho_{v}, w)+d(\rho_{w}, w).
\end{equation}
This ensures that \modelName is able to decode the pre-extracted textual label embeddings $w$ from the embeddings $\theta_a, \theta_v, \theta_w$.
The triplet loss $l_{ct}$ is defined as follows:
\begin{equation}\label{eq:composite_triplet}
  l_{ct}=t(\rho^{+}_w, \rho^{+}_a, \rho^{-}_a) + t(\rho^{+}_w, \rho^{+}_v, \rho^{-}_v),
\end{equation}
where $\rho^+$ and $\rho^-$ correspond to positive and negative examples respectively. $l_{ct}$ further encourages the decoded audio and visual features $\rho_a, \rho_v$ to be aligned with the textual features $\rho_w$ that were obtained using the same decoder (with shared weights).
The third component $l_w$ of $l_c$ is similar to the base triplet loss in \cref{eq:base_triplet} and compares the audio and visual embeddings $\theta_a, \theta_v$ to $\theta_w$:
\begin{equation}\label{eq:other_triplets}
\begin{aligned}
  l_w = t(\theta^+_w, \theta^+_a, \theta^-_a) + t(\theta^+_w, \theta^+_v, \theta^-_v) \\
        t(\theta^+_a, \theta^+_w, \theta^-_w) + t(\theta^+_v, \theta^+_w, \theta^-_w).
 \end{aligned}
\end{equation}

\mypara{Regularisation loss.}
The final component of our loss $l$ consists of regularisation loss terms which directly encourage the alignment of the audio and visual embeddings with the text embeddings while preserving the information from their respective input modality. For this, we add two reconstruction blocks $A_{rec}$ and $V_{rec}$, such that $\phi^{rec}_a=A_{rec}(\theta_a)$ and $\phi^{rec}_v=V_{rec}(\theta_v)$, $\phi^{rec}_a, \phi^{rec}_v \in \mathbb{R}^{k_f}$. $A_{rec}$ and $V_{rec}$ each consist of a linear projection layer followed by batch normalisation, ReLU, and dropout with dropout rate $r_{dec}$: 
\begin{equation}\label{eq:regularisation_loss}
\begin{aligned}
  l_r= d(\phi^{rec}_v, \phi_v)
  +d(\phi^{rec}_a, \phi_a)\\
  +d(\theta_{v}, \theta_{w})
  +d(\theta_{a}, \theta_{w}).
 \end{aligned}
\end{equation}

\section{Experiments}\label{sec:experiments}
We apply our \modelName model to audio-visual GZSL and ZSL for video classification. In this section, we first describe our proposed benchmark (\cref{sec:benchmark}). We discuss implementation details (\cref{sec:implementation_details}), and then ablate the choice of different model components and loss functions (\cref{sec:ablation_analysis}). Finally, we compare \modelName to state-of-the-art baseline methods for (G)ZSL (\cref{sec:results}), and provide a detailed qualitative analysis of the learnt multi-modal embeddings (\cref{sec:qualitative_results}).

\subsection{Audio-Visual GZSL Benchmark} \label{sec:benchmark}
In this section, we propose three benchmark datasets for audio-visual GZSL curated from the VGGSound~\cite{chen2020vggsound}, UCF101~\cite{soomro2012ucf101}, and ActivityNet~\cite{caba2015activitynet} datasets (summarised in \cref{tab:datasets_table})\footnote{VGGSound is covered by a Creative Commons license: \url{https://creativecommons.org/licenses/by/4.0/}, ActivityNet by the MIT license: \url{https://github.com/activitynet/ActivityNet/blob/master/LICENSE}.}, and introduce our training and evaluation protocol.

\mypara{Dataset statistics.} 
For our proposed audio-visual GZSL splits, we include classes contained in the Sports1M~\cite{karpathy2014sports} dataset only in our seen subsets to allow the use of feature extractors pretrained on Sports1M without leakage of information to unseen classes. 

Our GZSL splits for the three datasets consist of a training set (tr), a validation set which is divided into a subset with samples from seen classes (v(S)) and another one with unseen classes (v(U)). Finally, we provide a test set consisting of seen classes (ts(S)) and unseen classes (ts(U)). The training set and the seen validation subset share the same classes with a ratio of 0.9/0.1 with respect to the number of videos. The subsets $\{ \text{tr} \cup \text{v(U)} \cup \text{v(S)}\}$ and ts(S) share the same classes and were split to have a ratio of 0.9/0.1 with respect to the number of videos.

\begin{table}[t]
    \centering
    \setlength{\tabcolsep}{2pt}
    \renewcommand{\arraystretch}{1.2}
    \resizebox{\linewidth}{!}{%
    \begin{tabular}{l|c c | c c c c c c}
    \hline 
    Dataset & \multicolumn{2}{c|}{$\#$ classes} & \multicolumn{5}{c}{$\#$ videos} \\
    & all & tr / v(U) / ts(U) & tr & v (S) & v (U) & ts (S) & ts (U) \\ \hline
      VGGSound-GZSL & 276 & 138 / 69 / 69 & 70351 & 7817 & 3102 & 9032 & 3450\\
      UCF-GZSL & 51 & 30 / 12 / 9 & 3174 & 353 & 1467 & 555 & 1267\\
      ActivityNet-GZSL   & 200 & 99 / 51 / 50 & 9204 & 1023 & 4307 & 1615 & 4199\\
    
      \hline 
     \end{tabular}
    }
    \caption{Statistics for our VGGSound, UCF, and ActivityNet (G)ZSL datasets, showing the number ($\#$) of classes and videos in our splits (tr: train, v: validation, ts: test; S: seen, U: unseen).}
    \label{tab:datasets_table}
    \vspace{-1.2em}
\end{table}

\noindent \textit{VGGSound}~\cite{chen2020vggsound} is a large audio-visual dataset with 309 classes and over 200k videos. The videos can be grouped into the 9 categories \textit{animals, home, music, nature, people, sports, tools, vehicle}, and \textit{others}. For our VGGSound-GZSL split, we exclude videos from the \textit{others} category and all samples from v(U) and ts(U) that were used to train SeLaVi~\cite{asano2020labelling}, resulting in 93,752 videos in 276 classes. The 42 classes that overlap with the Sports1M dataset are only used as training classes for GZSL.

\noindent \textit{UCF101}~\cite{soomro2012ucf101} is a video action recognition dataset which consists of over 13k videos in 101 classes. We use the subset of UCF101 which contains audio information. This results in a total of 6,816 videos for 51 classes. 
Previous (visual-only) methods repeatedly split the dataset into random seen and unseen classes. The 30 classes contained in the Sports1M dataset are not selected as unseen classes.

\noindent \textit{ActivityNet}~\cite{caba2015activitynet} is an action recognition dataset with 20k videos in 200 classes of varying duration. Again, we propose the ActivityNet-GZSL split ensuring that the 99 classes contained in the Sports1M dataset are not selected as unseen classes. 

\begin{table*}[t]
\centering
\setlength{\tabcolsep}{4pt}
\renewcommand{\arraystretch}{1.2}
\resizebox{\linewidth}{!}{
\begin{tabular}{c|l|cccc|cccc|cccc}
\hline
Method type &Model & \multicolumn{4}{c}{VGGSound-GZSL} & \multicolumn{4}{c}{UCF-GZSL} & \multicolumn{4}{c}{ActivityNet-GZSL} \\
&  & S & U & HM & ZSL & S & U & HM & ZSL & S & U & HM & ZSL \\ \hline
\multirow{5}{*}{ZSL} 
& ALE~\cite{akata2015label}  & 0.28 &5.48 & 0.53 & 5.48 & 57.59 & 14.89 & 23.66 & 16.32 & 2.63  & 7.87  & 3.94 & 7.90\\
& SJE~\cite{akata2015evaluation}  & 48.33 &1.10 & 2.15 & 4.06 & 63.10 & 16.77 & 26.50 & 18.93 & 4.61 & 7.04  & 5.57 & 7.08\\
& DEVISE~\cite{frome2013devise}  & 36.22 & 1.07 & 2.08 & 5.59 & 55.59 & 14.94 & 23.56 & 16.09 & 3.45  & 8.53  & 4.91 & 8.53 \\
& APN~\cite{xu2020attribute}  & 7.48  & 3.88  & 5.11 & 4.49 & 28.46  & 16.16  & 20.61 & 16.44 &  9.84  & 5.76  & 7.27 & 6.34\\
 &f-VAEGAN-D2~\cite{xian2019f} & 12.77 & 0.95 & 1.77 & 1.91 & 17.29 & 8.47 & 11.37 & 11.11 & 4.36 & 2.14 & 2.87 & 2.40 \\
\hline
\multirow{2}{*}{Audio-visual} 
& CJME~\cite{parida2020coordinated}  & 8.69 &4.78 & 6.17 & 5.16 &26.04 &8.21 &12.48 &8.29 & 5.55 &4.75 & 5.12 &5.84\\
\multirow{2}{*}{ZSL} 
& AVGZSLNet~\cite{mazumder2021avgzslnet}  & 18.05 & 3.48 & 5.83 & 5.28 & 52.52 & 10.90 & 18.05 & 13.65 & 8.93 & 5.04 & 6.44 & 5.40 \\
 & \modelName & 14.90 &4.00&\textbf{6.31} & \textbf{6.00} & 51.53 & 18.43 & \textbf{27.15} & \textbf{20.01} & 24.86 & 8.02 & \textbf{12.13} & \textbf{9.13} \\
\hline
\end{tabular}
}
\caption{Evaluating our \modelName model and state-of-the-art audio-visual ZSL methods and adapted ZSL methods for GZSL and ZSL on the VGGSound, UCF, and ActivityNet (G)ZSL benchmarks. We report the mean class accuracy on the seen (S) and unseen (U) test classes, and their harmonic mean (HM) for GZSL performance. The ZSL performance is evaluated on the test subset from unseen classes.}
\label{tab:final_results}
\vspace{-1.1em}
\end{table*}

\mypara{Training and evaluation protocol.}
We introduce a unified training and evaluation protocol for our GZSL benchmarks. We follow this protocol to train and test all models, including \modelName and the baselines that we compare to.

We propose a two-stage training and evaluation protocol for GZSL.
In the first stage, we train the models on the training set (tr), using the subsets of seen validation classes (v(S)) and unseen validation classes (v(U)) to determine the GZSL parameters, for instance for calibrated stacking~\cite{chao2016empirical}.

In the second training stage, we re-train the models on the training (tr) and full validation set $\{ \text{v(S)} \cup \text{v(U)} \}$ using the GZSL parameters determined during the first training stage. Our final models are then evaluated on the test set $\{ \text{ts(S)} \cup \text{ts(U)} \}$. ts(S) contains samples from the same classes as the training classes with no overlap between training samples for the second stage and the test samples. In particular, there is no class overlap between \text{v(U)} and \text{ts(U)}.

\mypara{Evaluation metrics.} 
Following~\cite{xian2018zero}, we propose to evaluate all models using the mean class accuracy. For GZSL, we evaluate the models on the full test set $\{ \text{ts(S)} \cup \text{ts(U)} \}$, and report the averaged performance on the unseen (U) and seen (S) classes. Furthermore, we compute their harmonic mean $HM = \frac{2 U S}{U + S}$. We report the ZSL performance by evaluating only on the subset ts(U).

\subsection{Experimental Setting}\label{sec:implementation_details}
For each video, we use the self-supervised SeLaVi~\cite{asano2020labelling} framework pretrained on VGGSound~\cite{chen2020vggsound} to extract audio and visual features for each second in a video. In our VGGSound-GZSL split, there is no overlap between videos in the unseen test and unseen validation sets and videos that were used for pre-training SeLaVi. We average the per-second features extracted using SeLaVi prior to the two-layer MLP heads to obtain 512-dimensional per-video audio and visual features. 
We provide additional results for using features extracted from audio and video classification networks in the supplementary material.

All networks were optimised for GZSL performance (HM) and we do not train separate networks for GZSL and ZSL. 
The training for the first stage was done for $50$ epochs. We selected the number of training epochs for the second stage based on the GZSL performance on the validation set in the first stage.
To eliminate the bias that the ZSL methods have towards seen classes, we used calibrated stacking~\cite{chao2016empirical} on the interval $[0,3]$ with a step size of $0.2$. 
For \modelName, $k_{input}$ was set to $512$ and the size of the word2vec embedding, $k_{w2v}$, was set to $300$. We used dropout rates $r_{dec}$/$r_{enc}$/$r_{proj}$ of $0.5$/$0.2$/$0.3$ for UCF-GZSL, $0.1$/$0.2$/$0.2$ for Activity-GZSL, and $0.1$/$0$/$0$ for VGGSound-GZSL. 
The layer dimensions were set to $k_{f}=300$, $k_{fhidd}=512$, $k_{attnhidd}=64$, and $k_{proj}=64$. We used 3 heads for self-attention. The loss margin hyperparameter, $\mu$, was set to 1. We used a batchsize of 256 for UCF-GZSL and ActivityNet-GZSL, and 64 for VGGSound-GZSL.
We used the Adam optimiser~\cite{kingma2014adam} with an initial learning rate of $0.001$ which was reduced by a factor of $0.1$ when the GZSL performance plateaued with a patience of $3$ epochs.

\subsection{Comparing with the State of the Art}\label{sec:results}
\mypara{Compared methods.}
In our benchmark study, we include four image-based state-of-the-art methods and one generative method for (G)ZSL which we adapt to take audio-visual features as inputs.
For this, we concatenate the audio and visual features and use those as inputs instead of image features. Moreover, we compare to current state-of-the-art methods for audio-visual GZSL~\cite{parida2020coordinated,mazumder2021avgzslnet}.  
Here, we describe each of the methods that we compare to in more detail. 

\textbf{ALE}~\cite{akata2015label} learns a linear mapping between the input features and the ground-truth embeddings, such that the projection of the input features is close to the ground-truth embedding for the corresponding class. For this, it uses a weighted approximate ranking objective~\cite{usunier2009ranking}.
\textbf{SJE}~\cite{akata2015evaluation} computes the dot product between linearly mapped input features and the ground-truth embedding of all negative classes. The highest dot product for each example is chosen and then minimised.
\textbf{DEVISE}~\cite{frome2013devise} also computes the dot product between the output of a linear projection and the negative class embeddings and it minimises the sum of these dot products.
\textbf{APN}~\cite{xu2020attribute} is the current non-generative state-of-the-art method for image-based ZSL. APN is based on the assumption that the ground-truth embeddings contain visual class attributes. Prototypes are used to map the attributes from the ground-truth embeddings to relevant locations in the image. \textbf{f-VAEGAN-D2}~\cite{xian2019f} is a generative ZSL method which learns to generate synthetic features for unseen classes. Then, a classifier is trained on real examples from seen classes and synthetic examples from unseen classes.
\textbf{CJME}~\cite{parida2020coordinated} proposed the task of audio-visual GZSL for video classification on the AudioSetZSL dataset. It embeds audio, video and text into a joint embedding space and uses proximity in the embedding space to select the classification output at test time.
\textbf{AVGZSLNet}~\cite{mazumder2021avgzslnet} builds on \cite{parida2020coordinated} and is the current state-of-the-art method for audio-visual GZSL for video classification. One of the main strengths of this method is its use of triplet losses to leverage information from negative examples.

\mypara{Results.} We compare our \modelName framework to recent methods for (G)ZSL in \cref{tab:final_results} on the VGGSound-GZSL, UCF-GZSL, and ActivityNet-GZSL datasets.
\modelName obtains the best results on all three datasets. 
On VGGSound-GZSL, \modelName obtains a HM of 6.31\% for GZSL and a ZSL performance of 6.00\% compared to 6.17\% HM for CJME and a ZSL performance of 5.59\% for DEVISE. On the UCF-GZSL dataset, our \modelName model outperforms SJE for GZSL with a performance of 27.15\% compared to 26.50\%, and we obtain a stronger ZSL performance of 20.01\% compared to 18.93\%. 
On ActivityNet-GZSL, \modelName outperforms APN, with a GZSL performance of 12.13\% compared to 7.27\%. For ZSL, \modelName is stronger than DEVISE with a score of 9.13\% compared to 8.53\%. It can be observed that in some cases U is higher than S. This is due to the use of calibrated stacking~\cite{chao2016empirical} as described in~\cite{min2020domain}.

\begin{figure*}[t]
     \centering
     \begin{subfigure}[b]{0.27\textwidth}
         \centering
         \includegraphics[width=\textwidth]{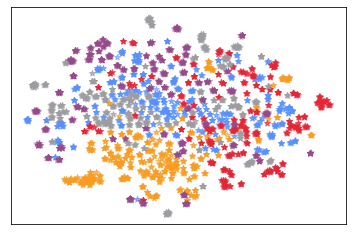}
         \caption{Input audio embeddings}
         \label{fig:audio_extraction}
     \end{subfigure}
     \begin{subfigure}[b]{0.27\textwidth}
         \centering
         \includegraphics[width=\textwidth]{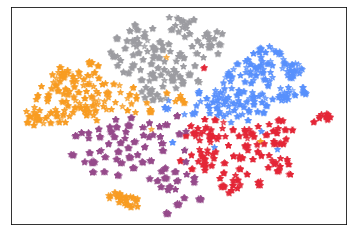}
         \caption{Input visual embeddings}
         \label{fig:video_extraction}
     \end{subfigure}
     \begin{subfigure}[b]{0.27\textwidth}
         \centering
         \includegraphics[width=\textwidth]{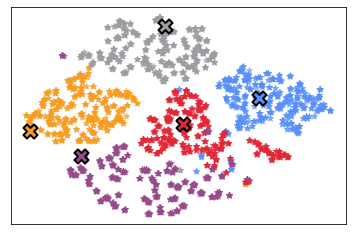}
         \caption{Learnt audio-visual embeddings}
         \label{fig:video_trained}
     \end{subfigure}
     \begin{subfigure}[b]{0.16\textwidth}
         \includegraphics[width=\textwidth,trim=0 -60 0 0]{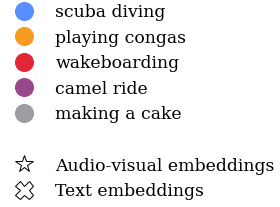}
         \label{fig:video_trained_label}
     \end{subfigure}
        \caption{t-SNE visualisation for three seen (\textit{scuba diving, playing congas, wakeboarding}) and two unseen (\textit{camel ride, making a cake}) test classes from ActivityNet-GZSL, showing embeddings extracted with SeLaVi~\cite{asano2020labelling} for (a) audio and (b) visual features. (c) Learnt audio-visual embeddings of our model. Projected textual class label embeddings are visualised with a cross with black boundary.}
        \label{fig:qualitative_results}
        \vspace{-0.9em}
\end{figure*}

\subsection{Qualitative Results}\label{sec:qualitative_results}
We present a qualitative analysis of the learnt multi-modal embeddings in~\cref{fig:qualitative_results}. The t-SNE visualisations~\cite{van2008visualizing} for a subset of ActivityNet-GZSL classes show the differences between the audio and visual input features and the learnt multi-modal embeddings. We provide additional qualitative results for VGGSound-GZSL and UCF-GZSL in the supplementary material. It can be seen in \cref{fig:audio_extraction} that the input audio features are not as well-separated and clustered as the visual features shown in \cref{fig:video_extraction}. However, the visual features also contain classes, such as \textit{playing congas} and \textit{scuba diving}, which are not clustered cleanly. It can be observed in \cref{fig:video_trained} that our model produces multi-modal features that improve over the clustering of the input embeddings for both, seen and unseen classes. For instance, the cluster separation between the seen class \textit{playing congas} and the unseen class \textit{making a cake} improves significantly, even though the unseen class is not used for training.

\subsection{Ablation Analysis}\label{sec:ablation_analysis}
\begin{table}[t]
\centering
\setlength{\tabcolsep}{4pt}
\renewcommand{\arraystretch}{1.2}
\resizebox{\linewidth}{!}{
\begin{tabular}{l|cc|cc|cc}
\hline
Model  & \multicolumn{2}{c}{VGGSound-GZSL}& \multicolumn{2}{c}{UCF-GZSL} & \multicolumn{2}{c}{ActivityNet-GZSL} \\
{} & HM&ZSL& HM &ZSL &HM&ZSL \\
\hline
Visual branch & 4.83 &4.06 & 20.92 &14.16 & 7.53 &6.49\\
Audio branch  & 3.84 &3.83& 11.78 & 10.78 & 4.19 & 4.06   \\
\modelName &\textbf{6.31}  &\textbf{6.00}& \textbf{27.15} & \textbf{20.01} &\textbf{12.13} &\textbf{9.13} \\
\hline
\end{tabular}
}
\caption{Influence of \textit{training} \modelName with different modalities for GZSL and ZSL on the VGGSound-GZSL, UCF-GZSL and ActivityNet-GZSL datasets measuring the harmonic mean (HM) for GZSL and the mean class accuracy for ZSL. Using both modalities yields the strongest GZSL and ZSL performances.}
\label{tab:branch_ablations_training}
\vspace{-1em}
\end{table}

Here, we analyse how different architectural choices and loss components for \modelName impact the performances on VGGSound-GZSL, ActivityNet-GZSL, and UCF-GZSL.

\mypara{Evaluating different modalities.} In \cref{tab:branch_ablations_training}, we compare our multi-modal \modelName model to training our architecture with only unimodal inputs. In this case, we remove the cross-modal attention block and train each unimodal branch in isolation. The visual branch obtains a better performance than the audio branch with a GZSL performance (HM) of 7.53\% vs. 4.19\%\ on the ActivityNet-GZSL dataset. A similar pattern can be observed for the ZSL performance with 6.49\% vs.\ 4.06\% for the visual and audio branch respectively. This trend is also exhibited on the UCF-GZSL and VGGSound-GZSL datasets, suggesting that the visual input features provide richer information about the video content than the audio inputs. Nevertheless, jointly training \modelName with both input modalities gives significant improvements over using each of them individually with a GZSL performance of 12.13\% and a ZSL performance of 9.13\% on the ActivityNet-GZSL dataset. This confirms that the complementary information from the audio and visual inputs is highly beneficial for GZSL and ZSL for video classification. We provide the S/U performances for \cref{tab:branch_ablations_training} in the supplementary material.

\begin{table}[t]
\centering
\setlength{\tabcolsep}{4pt}
\renewcommand{\arraystretch}{1.2}
\resizebox{\linewidth}{!}{
\begin{tabular}{l|cc|cc|cc}
\hline 
Model & \multicolumn{2}{c}{VGGSound-GZSL}& \multicolumn{2}{c}{UCF-GZSL} & \multicolumn{2}{c}{ActivityNet-GZSL} \\
{}& HM & ZSL & HM & ZSL & HM & ZSL \\
\hline
W/o x-att & 6.02& 4.81 & 26.82 & 18.37  & 6.50 & 5.64 \\
Visual with x-att & \textbf{6.63} & 4.78 & 27.11 & 17.22 & 9.50 & 6.89  \\ 
Audio with x-att  & 4.93 & 5.01 & 18.61 & 16.05 & 11.05 & 8.78  \\
\modelName & 6.31 & \textbf{6.00}& \textbf{27.15} & \textbf{20.01}& \textbf{12.13} &\textbf{9.13}\\
\hline
\end{tabular}
}
\caption{Using different components of \modelName for GZSL and ZSL on VGGSound-GZSL, UCF-GZSL and ActivityNet-GZSL. Audio (Visual) with x-att uses the visual (audio) modality only for the cross-attention. W/o x-att optimises each branch in isolation and their output predictions are averaged. x-att denotes cross-attention. 
}
\vspace{-1em}
\label{tab:branch_ablations_training_multimodal}
\end{table}

\mypara{Evaluating the cross-modal attention block.} Next, we investigate the effect of using our cross-modal attention block in \cref{tab:branch_ablations_training_multimodal}. 
To obtain results without using cross-attention (W/o x-att), each branch is optimised individually. For evaluation, we compute the distances between the outputs of both branches and $\theta_{w}$ for each class, and then average the distances computed by both branches. The GZSL and ZSL performances drop dramatically when not using the cross-attention block from 12.13\% and 9.13\% for \modelName to 6.50\% and 5.64\% for GZSL and ZSL scores respectively on the ActivityNet-GZSL dataset. The pattern is similar for VGGSound-GZSL and UCF-GZSL, confirming the importance of our cross-modal attention block for sharing information between the input modalities.

Furthermore, we compare optimising our full \modelName model to using only the visual (Visual with x-att) or only the audio branch (Audio with x-att) for training. Using only the visual branch entails removing $A_{rec}$ and $A_{proj}$ along with their associated losses from the audio branch but keeping the cross-attention. 
This experiment is repeated for the audio branch by removing the corresponding components from the visual branch. Jointly optimising both branches provides better results than using only one of the branches on ActivityNet-GZSL and UCF-GZSL. On ActivityNet-GZSL, we obtain a GZSL performances of 12.13\% compared to 11.05\% and 9.50\% for using only the audio and visual branches respectively. Interestingly, for the VGGSound-GZSL dataset, the Visual with x-att model yields a slightly stronger GZSL performance than our full \modelName model, with a HM of 6.63\% compared to 6.31\%. This is in line with the Audio branch performing worse than the Visual branch on VGGSound-GZSL (\cref{tab:branch_ablations_training}).
However, the joint optimisation of \modelName gives the best results. 

\begin{table}[t]
\centering
\setlength{\tabcolsep}{4pt}
\renewcommand{\arraystretch}{1.2}
\resizebox{\linewidth}{!}{
\begin{tabular}{l|cc|cc|cc}
\hline
Model output & \multicolumn{2}{c}{VGGSound-GZSL}& \multicolumn{2}{c}{UCF-GZSL} & \multicolumn{2}{c}{ActivityNet-GZSL}\\
{} & HM &ZSL& HM & ZSL & HM & ZSL \\ \hline
\modelName($\theta_a$) & 5.18 &4.87&  25.98 & 18.25 &\textbf{12.54}& \textbf{9.23}\\
\modelName($\theta_v$) & \textbf{6.31} & \textbf{6.00}&  \textbf{27.15}& \textbf{20.01} & 12.13 & 9.13 \\
\modelName($\theta_a, \theta_v$)& 5.90& 5.42&  25.78 & 19.30 & 12.17 & 8.95  \\

\modelName(min($\theta_a, \theta_v$)) &6.10 & 5.36& 25.86&18.39 &12.45 & 9.08\\
\hline
\end{tabular}
}
\caption{Influence of using the outputs of the audio and visual branches $\theta_a$ and $\theta_v$ separately, or using both jointly ($\theta_a$, $\theta_v$) for \textit{evaluation} on VGGSound-GZSL, UCF-GZSL and ActivityNet-GZSL. All models were trained with $\theta_a$ and $\theta_v$. }
\label{tab:branch_ablations}
 \vspace{-1em}
\end{table}

\mypara{Evaluating different modalities as output.} In \cref{tab:branch_ablations}, we investigate the effect of evaluating our full trained \modelName model using only the output features from the audio ($\theta_a$) or the visual ($\theta_v$) branch, or from both branches together (($\theta_a, \theta_v$) and $min$($\theta_a, \theta_v$)). For AVCA($\theta_a, \theta_v$), we compute the distance $|\theta_a - \theta_{w}|_2 + |\theta_v - \theta_{w}|_2$. AVCA($min$($\theta_a, \theta_v$)) uses the embedding from the modality that has the smallest distance to a word embedding. The class corresponding to the closest text embedding resembles the class prediction.

Using the visual branch gives the strongest performance on VGGSound-GZSL/UCF-GZSL with a HM of 6.31\%/27.15\% vs 5.18\%/25.98\% for the audio branch. On ActivityNet-GZSL, the audio branch yields slightly better results (HM of 12.54\% vs. 12.13\% for the visual branch). Both AVCA($\theta_a, \theta_v$) and AVCA($min$($\theta_a, \theta_v$)) obtain lower scores that $\theta_v$. The best results (highest averaged HM) across all three datasets are produced when using the visual branch only.
However, as the cross-attention block fuses the audio and visual modalities, both branches contain multi-modal information from both input modalities.

\mypara{Evaluating different loss functions.} Finally, we analyse the impact of using different loss functions for training \modelName on the GZSL and ZSL performance in \cref{tab:loss_ablations}. We observe that using our full loss $l$ provides the strongest GZSL results (HM) on the UCF-GZSL, VGGSound-GZSL, and ActivityNet-GZSL datasets by a large margin. 
On ActivityNet-GZSL, omitting $l_t$ for training our model ($l-l_t$) provides slightly stronger ZSL results than using our full loss $l$ with a mean class accuracy of 9.54\% compared to 9.13\%. However, the GZSL performance is significantly better when using $l$ with a HM of 12.13\% compared to 8.39\% when using $l-l_t$.
Our loss ablations confirm that our strong overall performance on all three datasets is only obtained when training with our full proposed loss function.

\begin{table}[t]
\centering
\setlength{\tabcolsep}{5pt}
\renewcommand{\arraystretch}{1.1}
\resizebox{\linewidth}{!}{
\begin{tabular}{l|cc|cc|cc}
\hline
Model  & \multicolumn{2}{c}{VGGSound-GZSL} & \multicolumn{2}{c}{UCF-GZSL} & \multicolumn{2}{c}{ActivityNet-GZSL} \\
{}& HM &ZSL& HM & ZSL & HM & ZSL \\
\hline
$l$-$l_{t}$ & 5.06 & 4.84 & 18.51& 19.17& 8.39 &\textbf{9.54} \\
$l$-$l_{rec}$  & 5.92 & 5.22 &24.32  &17.20 & 9.59 &6.93 \\
$l$-$l_{ct}$  & 6.31& 4.87&17.88&17.51 & 11.20& 8.99\\
$l$-$l_{w}$  & 5.18 & 4.93  & 20.75& 16.41  & 9.08 &8.00 \\
$l$-$l_{r}$ & 6.24  & 4.43   & 21.31&14.02 & 11.14&7.94 \\
$l$       & \textbf{6.31}& \textbf{6.00}     &\textbf{27.15} &\textbf{20.01}   &\textbf{12.13} &9.13 \\
\hline

\end{tabular}
}
\caption{Comparing training \modelName with our full loss function~$l$ to removing individual components $l_t$, $l_{rec}$, $l_{ct}$, $l_{w}$, or $l_r$, on the GZSL and ZSL performance on the VGGSound-GZSL, UCF-GZSL and ActivityNet-GZSL datasets.}
\label{tab:loss_ablations}
 \vspace{-0.9em}
\end{table}

\subsection{Limitations and Discussion}
Our proposed GZSL benchmark datasets pose an extremely challenging setting, since the underlying datasets span a wide variety of classes (\eg including \textit{wakeboarding} and \textit{making a cake} for the ActivityNet dataset). 
Our \modelName leverages the varied audio-visual input information effectively, resulting in more robust GZSL performance than the related methods.
However, \modelName uses temporally averaged audio-visual input information, and hence does not consider fine semantic details.
%
Furthermore, our model relies on multi-modal input data and cannot be used when only one modality is available.

\section{Conclusion}
We introduced three new benchmarks for audio-visual (generalised) zero-shot learning for video classification on the VGGSound, UCF, and ActivityNet datasets. We proposed a framework for (G)ZSL from audio-visual data which learns to align the audio-visual embeddings with textual label embeddings. Furthermore, we provided baseline performances for seven (G)ZSL methods, and show that our model outperforms them for GZSL and ZSL on our new benchmarks. Finally, we provided a qualitative analysis of the learnt multi-modal embeddings. We hope that our proposed benchmarks will enable and encourage further research into audio-visual zero-shot learning.

\section*{Acknowledgements} 
This work was supported by BMBF FKZ: 01IS18039A, DFG: SFB 1233 TP 17, project number: 276693517, by the ERC (853489 - DEXIM), and by EXC number 2064/1 – Project number 390727645. The authors thank the International Max Planck Research School for Intelligent Systems (IMPRS-IS) for supporting O.-B. Mercea. The authors would like to thank M. Mancini for helpful suggestions and feedback.

\newpage
\section*{Supplementary Material: Audio-visual Generalised Zero-shot Learning with Cross-modal Attention and Language}


\appendix
\noindent In this supplementary material, we include additional qualitative results (\cref{sec:supp}) and quantitative results (\cref{sec:quant}) for our proposed audio-visual (G)ZSL framework.

\section{Additional Qualitative Results}\label{sec:supp}
\begin{figure}[b]
     \vspace{-1em}
     \begin{subfigure}[b]{0.22\textwidth}
         \centering
         \includegraphics[width=\textwidth]{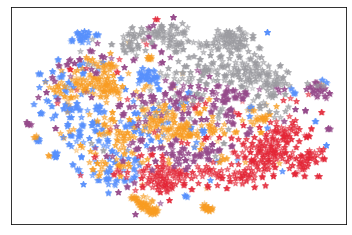}
         \caption{Input audio embeddings}
         \label{fig:audio_extraction_vgg}
     \end{subfigure}
     \begin{subfigure}[b]{0.22\textwidth}
         \centering
         \includegraphics[width=\textwidth]{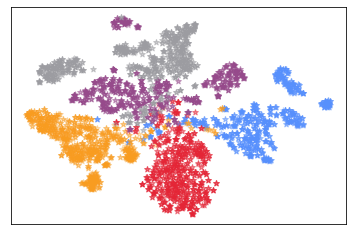}
         \caption{Input visual embeddings}
         \label{fig:video_extraction_vgg}
     \end{subfigure}
     \begin{subfigure}[b]{0.22\textwidth}
         \centering
         \includegraphics[width=\textwidth]{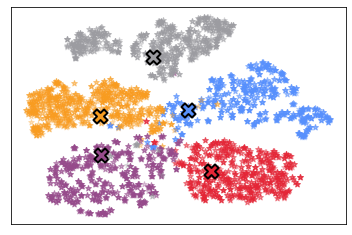}
         \caption{Learnt audio-visual embeddings}
         \label{fig:video_trained_vgg}
     \end{subfigure}
    \hspace{1.2em}
     \begin{subfigure}[b]{0.17\textwidth}
         \centering
         \includegraphics[width=\textwidth,trim= 0  -5  0 0 , clip]{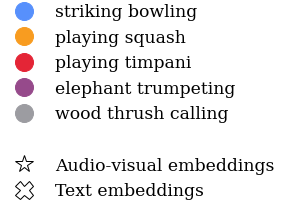}
         \label{fig:video_trained_label_vgg}
     \end{subfigure}
         \caption{t-SNE visualisation for three seen (\textit{striking bowling, playing squash, playing timpani}) and two unseen (\textit{elephant trumpeting, wood thrush calling}) test classes from the VGGSound-GZSL dataset, showing (a) audio and (b) visual features extracted with SeLaVi~\cite{asano2020labelling}, and (c) learnt audio-visual embeddings of our model. Textual class label embeddings are visualised with a cross.}
        \label{fig:qualitative_results_vgg}
\end{figure}

\begin{figure}[t]
     \vspace{-0.5em}
     \begin{subfigure}[b]{0.22\textwidth}
         \includegraphics[width=\textwidth]{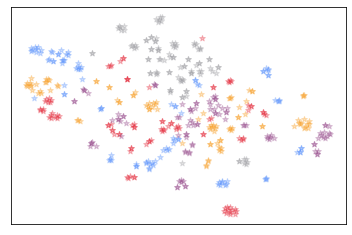}
         \caption{Input audio embeddings}
         \label{fig:audio_extraction_ucf}
     \end{subfigure}
     \begin{subfigure}[b]{0.22\textwidth}
         \includegraphics[width=\textwidth]{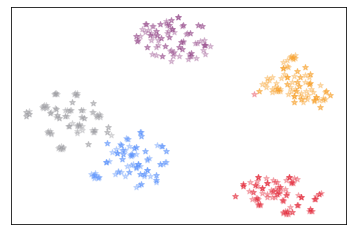}
         \caption{Input visual embeddings}
         \label{fig:video_extraction_ucf}
     \end{subfigure}
     \begin{subfigure}[b]{0.22\textwidth}
         \includegraphics[width=\textwidth]{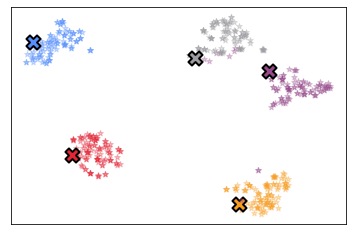}
         \caption{Learnt audio-visual embeddings}
         \label{fig:video_trained_ucf}
     \end{subfigure}
     \hspace{1.2em}
     \begin{subfigure}[b]{0.17\textwidth}
         \centering
         \includegraphics[width=\textwidth,trim=0 0 0 5, clip]{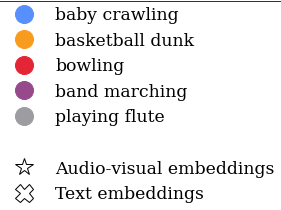}
         \label{fig:audiovisual_ucf}
     \end{subfigure}
        \caption{t-SNE visualisation for three seen (\textit{baby crawling, basketball dunk, bowling}) and two unseen (\textit{band marching, playing flute}) test classes from the UCF-GZSL dataset, showing (a) audio and (b) visual features extracted with SeLaVi~\cite{asano2020labelling}, and (c) learnt audio-visual embeddings of our model. Textual class label embeddings are visualised with a cross.}
        \label{fig:qualitative_results_ucf}
\end{figure}

We provide additional qualitative results for our proposed \modelName model for the tasks of audio-visual GZSL and ZSL.  
We present t-SNE visualisations for the learnt audio-visual embeddings on the VGGSound-GZSL and UCF-GZSL datasets in \cref{fig:qualitative_results_vgg} and \cref{fig:qualitative_results_ucf}. 

In \cref{fig:audio_extraction_vgg}, we can observe that the input audio features do not demonstrate a clear separation between the visualised classes for the VGGSound-GZSL dataset. The visual features exhibit a better clustering as can been seen in \cref{fig:video_extraction_vgg}. However, the visual features also include classes, such as \textit{elephant trumpeting} and \textit{wood thrush calling}, that are not clustered cleanly. Our \modelName model outputs multi-modal features that improve the clustering for both, seen and unseen classes (\cref{fig:video_trained_vgg}). The learnt features for the two unseen classes \textit{elephant trumpeting} and \textit{wood thrush calling} are clustered and well-separated as opposed to the input features. This is impressive, since both classes were not included in the training set.

Similarly, for the UCF-GZSL dataset, we can observe in \cref{fig:audio_extraction_ucf} that the input audio features are not grouped according to classes. In contrast, the visual input embeddings mostly exhibit a clear clustering of different classes. However, the classes \textit{baby crawling} and \textit{playing flute} are not well-separated as can be seen in \cref{fig:video_extraction_ucf}. This improves through learning, since the learnt audio-visual features in \cref{fig:video_trained_ucf} show a clear divide between those two classes. In addition to that, the output embeddings for the unseen classes \textit{band marching} and \textit{playing flute} are overwhelmingly clustered well, too.

To summarise, our model learns to cluster both seen and unseen classes for different datasets by transferring information from the training data to unseen classes at test time.

\section{Additional Quantitative Results}\label{sec:quant}
\begin{table*}[t]
\centering
\setlength{\tabcolsep}{4pt}
\renewcommand{\arraystretch}{1.2}
 \resizebox{\linewidth}{!}{
 \begin{tabular}{c | l|cccc|cccc|cccc}
 \hline
 Method type & Model & \multicolumn{4}{c}{VGGSound-GZSL$^{cls}$} & \multicolumn{4}{c}{UCF-GZSL$^{cls}$} & \multicolumn{4}{c}{ActivityNet-GZSL$^{cls}$} \\
 &  & S & U & HM & ZSL & S & U & HM & ZSL & S & U & HM & ZSL \\ \hline
 \multirow{4}{*}{ZSL} 
 & ALE~\cite{akata2015label}  &26.13&1.72&3.23&4.97&45.42&29.09&35.47&32.30&0.89 &6.16&1.55&6.16  \\
 & SJE~\cite{akata2015evaluation}  & 16.94&2.72&4.69&3.22&19.39&32.47&24.28&32.47& 37.92&1.22&2.35&4.35\\
 & DEVISE~\cite{frome2013devise}  &29.96&1.94& 3.64&4.72&29.58&34.80&31.98&35.48&0.17&5.84&0.33&5.84 \\
 & APN~\cite{xu2020attribute} & 6.46&6.13&6.29&6.50&13.54&28.44&18.35&29.69&3.79&3.39&3.58&3.97  \\
 \hline
 Audio-visual & CJME~\cite{parida2020coordinated} &10.86&2.22&3.68&3.72&33.89&24.82&28.65&29.01&10.75&5.55&7.32&6.29  \\
 ZSL & AVGZSLNet~\cite{mazumder2021avgzslnet} &15.02&3.19&5.26&4.81&74.79&24.15&36.51&31.51&13.70&5.96&8.30&6.39 \\
 & AVCA &12.63&6.19&\textbf{8.31}&\textbf{6.91}&63.15&30.72&\textbf{41.34}&\textbf{37.72}&16.77&7.04&\textbf{9.92}&\textbf{7.58}  \\
 \hline
 \end{tabular}
 }
 \caption{Evaluating \modelName and state-of-the-art (G)ZSL methods for audio-visual GZSL and ZSL on the VGGSound, UCF, and ActivityNet (G)ZSL$^{cls}$ benchmarks using features extracted from audio/video classification networks. We report the mean class accuracy on the seen (S) and unseen (U) test classes, and their harmonic mean (HM) for GZSL performance. The ZSL performance is evaluated on the test subset of samples from unseen classes.}
 \label{tab:final_results_supervised}
 \end{table*}

In this section, we provide additional quantitative results obtained with our \modelName. We present results for training and evaluating our \modelName model with a different set of input features in \cref{sec:supp_supervised}. In particular, we use features extracted from networks that were pretrained for audio and video classification. We perform an additional ablation study that gradually transforms \modelName into AVGZSLNet~\cite{mazumder2021avgzslnet} in \cref{sec:supp_avgzsl}. Complete results that include the U and S performance for Table~3 in the main paper are provided in \cref{sec:supp_full_table}. Finally, we give details about the number of parameters and GFLOPS required for training our \modelName model in \cref{sec:supp_params}

\subsection{Using features extracted audio/video classification networks}\label{sec:supp_supervised}

We additionally trained and tested our model and the baseline models using features extracted from audio and video classification networks (instead of the SeLaVi~\cite{asano2020labelling} features used in the main paper). In particular, the visual features were extracted with C3D~\cite{tran2015learning}, pretrained for video classification on Sports1M~\cite{karpathy2014sports}. The audio features were extracted with VGGish~\cite{hershey2017cnn}, pretrained for audio classification on Youtube-8M~\cite{abu2016youtube}. We averaged the extracted features across time, resulting in a 4096-dimensional visual feature and a 128-dimensional audio feature for each video.

 However, to use the audio features extracted from a network that was pretrained on Youtube-8M, we removed the test unseen classes from the VGGSound-GZSL, UCF-GZSL, and ActivityNet-GZSL datasets that had an overlap with Youtube-8M. 
 This resulted in slightly different dataset splits (VGGSound-GZSL$^{cls}$, UCF-GZSL$^{cls}$, and ActivityNet-GZSL$^{cls}$) detailed in~\cref{tab:datasets_table_vggish}.
 
 \begin{table}[t]
    \centering
    \setlength{\tabcolsep}{2pt}
    \renewcommand{\arraystretch}{1.2}
    \resizebox{\linewidth}{!}{%
    \begin{tabular}{l|c c c c | c}
    \hline 
    Dataset & \multicolumn{4}{c|}{$\#$ classes} & \multicolumn{1}{c}{$\#$ videos} \\
    & all & tr & v(U) & ts(U) & ts(U)\\ \hline
      VGGSound-GZSL$^{cls}$ & 271 & 138 & 69 & 64 & 3200\\
      UCF-GZSL$^{cls}$ & 48 & 30 & 12 & 6 & 845 \\
      ActivityNet-GZSL$^{cls}$   & 198 & 99 & 51 & 48 & 4052 \\
      \hline 
     \end{tabular}
    }
    \caption{Statistics for our VGGSound, UCF, and ActivityNet (G)ZSL$^{cls}$ datasets, showing the number ($\#$) of classes in our splits (tr: train, v: validation, ts: test; S: seen, U: unseen). $^{cls}$ indicates the dataset splits that allow to use VGGish features pre-trained on YouTube-8M. The full details about the dataset splits can be found at \url{https://github.com/ExplainableML/AVCA-GZSL}.}
    \label{tab:datasets_table_vggish}
\end{table}

We provide results for training and evaluating our \modelName and the baselines using audio and video classification features in \cref{tab:final_results_supervised}.
\modelName outperforms all the baselines on all three datasets. On VGGSound-GZSL$^{cls}$, ACVA obtains a HM of 8.31\% and ZSL of 6.91\% compared to a HM of 6.29\% for APN and a ZSL performance of 6.50\% for APN. On UCF-GZSL$^{cls}$, AVCA obtains a HM of 41.34\% and a ZSL of 37.72\% compared to a HM of 36.51\% for AVGZSLNet and a ZSL performance of 35.48\% for DEVISE. On ActivityNet-GZSL$^{cls}$, AVCA outperforms AVGZSLNet with a HM of 9.92\% compared to 8.30\% and a ZSL of 7.58\% compared to 6.39\% for AVGZSLNet. These results show that AVCA outperforms the other competitors also when using audio and video classification features, proving again that our cross-attention mechanism and training objective provide a boost in performance.

\subsection{Ablating \modelName in relation to AVGZSLNet}\label{sec:supp_avgzsl}
\begin{table}[t]
\centering
\setlength{\tabcolsep}{4pt}
\renewcommand{\arraystretch}{1.2}
\resizebox{\linewidth}{!}{
\begin{tabular}{l|cc|cc|cc}
\hline 
Model & \multicolumn{2}{c}{VGGSound-GZSL}& \multicolumn{2}{c}{UCF-GZSL} & \multicolumn{2}{c}{ActivityNet-GZSL} \\
{}& HM&ZSL& HM & ZLS &HM & ZSL \\
\hline
AVGZSLNet~\cite{mazumder2021avgzslnet} & 5.83 & 5.28 & 18.05 & 13.65 & 6.44 & 5.40\\
W/o x-att & 6.02& 4.81 & 26.82 &18.37  & 6.50&5.64 \\
W x-att with $l_c$ loss & 4.88 & 4.55 & 19.38& 12.95& 11.58 & 8.40 \\
\modelName & \textbf{6.31} & \textbf{6.00} & \textbf{27.15} & \textbf{20.01} & \textbf{12.13} & \textbf{9.13}\\
\hline
\end{tabular}
}
\caption{Ablation that gradually transforms our \modelName model into AVGZSLNet~\cite{mazumder2021avgzslnet}. W/o x-att optimises each branch in isolation and their output predictions are averaged. x-att denotes cross-attention. $l_c$ loss is the loss function used to train AVGZSLNet.
}
\label{tab:incremental_ablation}
\end{table}

We additionally perform an ablation study that gradually transforms the \modelName model into AVGZSLNet~\cite{mazumder2021avgzslnet} in \cref{tab:incremental_ablation}. We show how our model components influence the (G)ZSL performance, resulting in our \modelName model that outperforms AVGZSLNet on all three datasets. For this ablation, we use the SeLaVi~\cite{asano2020labelling} features and the same setup as in the main paper. 
W/o x-att corresponds to AVGZSLNet trained with our loss function (without our cross-attention). It can be observed that W/o x-att provides improvements on UCF-GZSL, with a HM of 26.82\% compared to 18.05\% and a ZSL performance of 18.37\% compared to 13.65\%. W x-att with $l_c$ loss corresponds to AVGZSLNet with cross-attention and  with the loss function proposed for AVGZSLNet. In this case, it can be observed that the cross-attention improves the results over AVGZSLNet with a HM of 11.58\% compared to 6.44\% and ZSL performance of 8.40\% compared to 5.40\% on ActivityNet-GZSL. These improvements can also be observed on the other datasets, showing that our novel loss and our cross-attention mechanism improve the performance over AVGZSLNet.

\subsection{Extended results for training AVCA with different modalities}\label{sec:supp_full_table}
\begin{table}[t]
\centering
\setlength{\tabcolsep}{4pt}
\renewcommand{\arraystretch}{1.2}
\resizebox{\linewidth}{!}{
\begin{tabular}{l|ccc|ccc|ccc}
\hline
Model  & \multicolumn{3}{c}{VGGSound-GZSL}& \multicolumn{3}{c}{UCF-GZSL} & \multicolumn{3}{c}{ActivityNet-GZSL} \\
{} & S & U & HM & S & U & HM & S & U & HM \\
\hline
Visual branch &7.02&3.68& 4.83  &50.18&13.21& 20.92  &11.80&5.53& 7.53 \\
Audio branch &7.74&2.55 & 3.84 & 12.99 &10.78&11.78  &4.56&3.87& 4.19    \\
\modelName &14.90&4.00 &\textbf{6.31}  &51.53 &18.43&\textbf{27.15} & 24.86&8.02&\textbf{12.13} \\
\hline
\end{tabular}
}
\caption{Influence of \textit{training} \modelName with different modalities for GZSL on the VGGSound-GZSL, UCF-GZSL and ActivityNet-GZSL datasets measuring the GZSL performance on seen (S) and unseen (U) test classes and their harmonic mean (HM). Using both modalities yields the strongest GZSL performances.}
\label{tab:branch_ablations_training_full}
\vspace{-1em}
\end{table}

In this section, we extend the ablation study that uses different modalities for training (Table~3 in the main paper) by adding the performance on the seen (S) and unseen (U) test classes for all the datasets in \cref{tab:branch_ablations_training_full}. 

On all three datasets it can be observed that there is an increase in both seen and unseen performance when using \modelName compared to using the Visual branch or the Audio branch. On VGGSound-GZSL, we can observe that the S performance for AVCA is 14.90\% compared to 7.74\% for the Visual branch.  The U performance on VGGSound-GZSL is also stronger for \modelName than for the Visual branch, with a score of 4.00\% compared to 3.68\%. On the UCF-GZSL dataset, the S performance increases only slightly, from 50.18\% for the Visual branch to 51.53\% for AVCA. However, there is a significant increase in the U performance, from 13.21\% for the Visual branch to 18.43\% for AVCA. Finally, on ActivityNet-GZSL, AVCA yields a S score of 24.86\% compared to 11.80\% for the Visual branch. The U performance increases from 5.53\% for the Visual branch to 8.02\% for AVCA. These results show that the S/U performance increases significantly when using AVCA compared to the Visual branch or the Audio branch, leading to better HM/ZSL performances.

\subsection{Number of parameters in \modelName.}\label{sec:supp_params}
\modelName contains 1.69M parameters in total, which is comparable to the 1.32M parameters used in AVGZSLNet~\cite{mazumder2021avgzslnet}. ALE/SJE/DEVISE are significantly smaller with only 307.2k parameters. \modelName has a computational complexity of 2.36 GFLOPS, while AVGZSLNet has a computational complexity of 1.38 GFLOPS. Again, the fewest GFLOPS are required for ALE/SJE/DEVISE which have a computational complexity of 0.32 GFLOPS. These statistics show that \modelName is comparable to AVGZSLNet while providing significantly better results on all three datasets.

{\small
\bibliographystyle{ieee_fullname}
\bibliography{egbib}
}

\end{document}